\documentclass[final]{cvpr}

\usepackage{times}
\usepackage{epsfig}
\usepackage{graphicx}
\usepackage{amsmath}
\usepackage{amssymb}

\usepackage[pagebackref=true,breaklinks=true,colorlinks,bookmarks=false]{hyperref}

\begin{document}


\title{Machine Learning approaches to do size based reasoning on Retail Shelf objects to classify product variants} 

\author{Muktabh Mayank Srivastava \hspace{12pt} Pratyush Kumar \hspace{12pt}  \\
\href{https://paralleldots.com}{ParallelDots, Inc.}\\}

\maketitle

\begin{abstract}
There has been a surge in the number of Machine Learning methods to analyze products kept on retail shelves images. Deep learning based computer vision methods can be used to detect products on retail shelves and then classify them. However, there are different sized variants of products which look exactly the same visually and the method to differentiate them is to look at their relative sizes with other products on shelves. This makes the process of deciphering the sized based variants from each other using computer vision algorithms alone impractical. In this work, we propose methods to ascertain the size variant of the product as a downstream task to an object detector which extracts products from shelf and a classifier which determines product brand. Product variant determination is the task which assigns a product variant to products of a brand based on the size of bounding boxes and brands predicted by classifier.  While gradient boosting based methods work well for products whose facings are clear and distinct, a noise accommodating Neural Network method is proposed for cases where the products are stacked irregularly. 
\end{abstract}

\section{Introduction}

The recent surge in the automation of retail and consumer industries has demanded to work on the intersection of different horizons of AI. One of the primary needs of automation in the retail industry is to detect and recognize products placed on shelves in stores. In the last few years, deep convolutional neural networks (CNNs) \cite{Simonyan2014VeryRecognition} have outperformed the state of the art in many visual recognition tasks. CNNs have been responsible for the phenomenal advancements in tasks like object classification \cite{KrizhevskyImageNetNetworks}, object detection \cite{he2017mask} etc., and the continuous improvements to CNN architectures are bringing further progress. For retail shelves, CNN based architectures have been used to detect products on shelves \cite{varadarajan2020benchmark} and then classify them \cite{MayankSrivastavaBagClassification}. Due to the high number of possible objects and their placements and lack of training data, a generic object detection algorithm is preferred to extract products from retail shelf images and the extracted products are then classified by a separate CNN classifier. \\*

However, shelf analysis might require not just detection and identification of products, but also deciphering which size variants of a product are placed on shelf. For example, a 100 Gram and a 200 Gram packet of a cereal brand B can be placed on shelf and it might be required to not just identify the cereal brand B, but also to identify size variants of the cereal brand placed on the shelf. These products are often visually similar apart from their relative sizes on shelf and product quantity mentioned on them. Mentioned product quantity is non-readable in full sized shelf images and hence the relative size of bounding boxes from Object Detection algorithm is the only method to differentiate these size variants in the image.  While we get different sized bounding boxes from CNN Object Detection, we pass input patches of the same dimensions to the CNN classifier and hence size based inference is impractical from the normal classifier algorithm. We thus use the products predicted by CNN classifier and sizes of bounding boxes from Object Detection method as input to a third step to do size reasoning and differentiate size variants of products. Two methods are proposed for the task, using features available in different ways.

\begin{figure*}[hbt!]
  \centering
  \includegraphics[width=\textwidth]{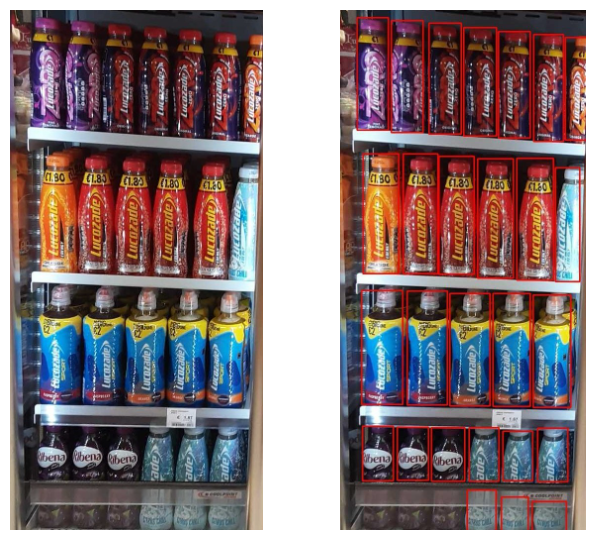}
  \caption{An image of a shelf in a store (left side). The image (right side) is passed through CNN Object detector which extracts products. Red Boxes show the predicted bounding boxes of products. These bounding boxes are then passed through a CNN classifier one at a time to determine brands and even later through a size based reasoning workflow to determine product size variants}
  \label{fig:shelf}
\end{figure*}

Many features from Object Detection and classification pipelines can be used in downstream tasks of product size variant identification e..g. Product brand, frontal area, aspect ratio. RGBD images if taken in place of RGB images also have depth information available, but that is outside the scope of this work where we work on RGB images only. There can be an alternative approach to somehow incorporate size variant information in the training of Computer Vision Object Detector and classifier, however, our focus has been to build a downstream method which can be used with any CNN object detector and classification algorithm. \\*

So, to sum up our size based reasoning model is a downstream task after A. CNN object detector is run on image to extract products and B. CNN classifier classifies product brands. Experiments in this work assume that the tasks A and B have been performed with 100\% accuracy and then do size based reasoning, any errors in upstream tasks would effect results downstream. Thus, size based inference is done on product bounding boxes from object detector and inferred product classes [deciphered brands] from CNN classifier. To establish some vocabulary for further discussion, we would highlight the following terms we will use henceforth :

\begin{enumerate}

\item Boxes are the rectangular boxes that object detectors like \cite{varadarajan2020benchmark} marks on images to extract different products from shelves.  
\item The inferred brands from the CNN classifier are called “Groups”. Since sized based reasoning is a further classification task into one of the products’ size variants, the output of CNN classifier can thus denote a group of sized variant classes and hence the term is used. 
\item Each bounding box belonging to a specific group [that is combining output of Computer Vision object detector and CNN image classifier] can be further classified into a size variant of the group [referred to as product brand in discussion till now]. We solve the problem of size based reasoning as a classification task and hence the size based variants are referred to as “classes” henceforth.

\end{enumerate}

The features available for downstream task are used in different ways in the two methods we propose. While the first method tries to coalesce the variable length features into a fixed sized feature vector which can be passed through a XGBOOST classifier. In the second method, features are modelled as mixture of gaussians and probabilities of belonging to different gaussians deduced are used to feed to a Neural Network classifier which can work on a variable length input.

\begin{figure*}[tb]
  \centering
  \includegraphics[width=0.5\textwidth,height=0.3\textheight]{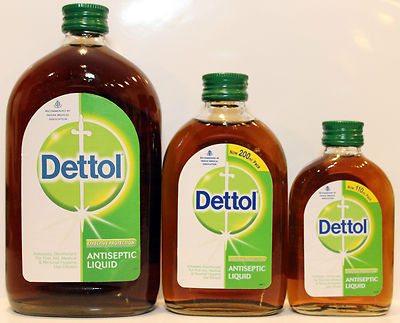}
  \caption{Image showing size variants of Dettol Antiseptic Liquid. In this work's terminology, all three bottles would belong to the same group but different classes. A CNN classifier can distinguish groups from each other, but cannot differentiate classes within groups due to its fixed input size. Reader can visualize this limitation of CNN classifier by thinking of how similar the size variants would look when their images are looked at one at a time in full vision field. This is the reason why the group level output of CNN classifier and bounding boxes from object detector need to be combined for size based reasoning of product variants. }
  \label{fig:sizeproducts}
\end{figure*}

\section{Related Work}
State-of-the-art object recognition methods rely heavily on deep CNNs, as further discussed below. Because of the limitations of CNN-based methods, hybrid methods, which combine CNNs prediction with background knowledge and knowledge-based reasoning, have been proposed in this paper.

\subsection{CNN}
CNNs have been extensively used to classify images \cite{KrizhevskyImageNetNetworks} \cite{Simonyan2014VeryRecognition} and detect objects within them \cite{he2017mask}. The new trend in detecting and recognizing objects on retail shelf is to first extract all objects from shelf using an object detection model \cite{varadarajan2020benchmark} and then classify each extracted box separately using a CNN classifier into a product brand \cite{MayankSrivastavaBagClassification}. In this work, our machine learning models are downstream from both these steps to further classify the product brand into one of the size based variants.

\begin{figure}[tb]
  \centering
  \includegraphics[width=0.4\textwidth]{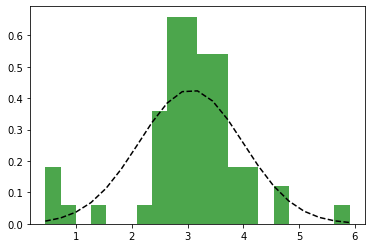}
  \caption{The plot shows distribution of aspect ratio for one of a group. The data roughly follow normal distribution with some extreme noises.}
  \label{fig:noise}
\end{figure}

\subsection{Gaussian Mixture Models and Neural Network for Classification}
A Gaussian mixture model is a probabilistic model that assumes all the data points are generated from a mixture of a finite number of Gaussian distributions with unknown parameters. A Gaussian mixture model is parameterized by two types of values, the mixture component weights and the component means and variances/covariances. 
For a Gaussian mixture model with K components, the $k^{th}$ component has a mean of $\mu_k$ and variance of $\sigma_k$ for the univariate case and a mean of $\vec{\mu}_k$ and covariance matrix of $\Sigma_k$ for the multivariate case. The mixture component weights are defined as $\phi_k$  for component $C_k$, with the constraint that $\sum_{i=1}^K\phi_i = 1$ so that the total probability distribution normalizes to 1.\\*

The Gaussian distribution for multivariate is, 
\begin{equation}
\mathcal{N}(\vec{x} | \vec{\mu}, \Sigma) = \frac{1}{(2\pi)^{K/2}}\frac{1}{|\Sigma|^{1/2}}exp\{-\frac{1}{2}(\vec{x}-\vec{\mu})^T\Sigma^{-1}(\vec{x}-\vec{\mu})\}
\end{equation}

Recently, GMM has been used for some types of classification problems like anomaly detection \cite{ZongDEEPDETECTION} where we have the prior knowledge of data distribution forming clusters with different mean and standard deviation. Mixture models in general do not require knowledge of which subpopulation a data point belongs to, allowing the model to learn the subpopulations automatically. Since subpopulation assignment is not known, this constitutes a form of unsupervised learning. GMM is harder to achieve in case the number of subpopulations is unknown. \cite{Wan2019AClassification} provided a novel approach to automatically find the number of mixture components in a dataset based on the separability criterion in order to separate the Gaussian models as much as possible. However, if we have prior information of the number of components or clusters in a dataset, the problem doesn't require making any guesses about the number of subpopulations. GMM has been proven to handle noisy data more efficiently compared to conventional machine learning algorithms. \cite{Hou2008RobustData} proposed a robust estimation gaussian mixture from noisy data taking into account the uncertainty associated with each data point, makes no assumptions about the structure of the covariance matrices and is able to automatically determine the number of the gaussian mixture components. 

\subsection{Boosting}
Boosting is a machine-learning method which is widely used in many classification and regression problems. It is a way to get performance of a strong learner by aggregating many weak learners. To convert weak learner to strong learner, it combines the prediction of each weak learner using methods like average/ weighted average. To find weak rule, it applies  base learning (ML) algorithms with a different distribution. Each time base learning algorithm is applied, the base learning algorithm generates a new weak prediction rule, and after many rounds, the boosting algorithm must combine these weak rules into a single strong prediction rule that, hopefully, will be much more accurate than any one of the weak rules. \\*
XGBoost \cite{Chen2016XGBoost:System}, is a scalable tree boosting system that provides state-of-the-art results on many classification and regression problems. In prediction problems having small-to-medium structured/tabular data, decision tree based algorithms perform very well. It is an ensemble tree method that applies the principle of boosting weak learners using the gradient descent architecture.

\begin{figure*}[hbt!]
  \centering
  \includegraphics[width=0.8\textwidth]{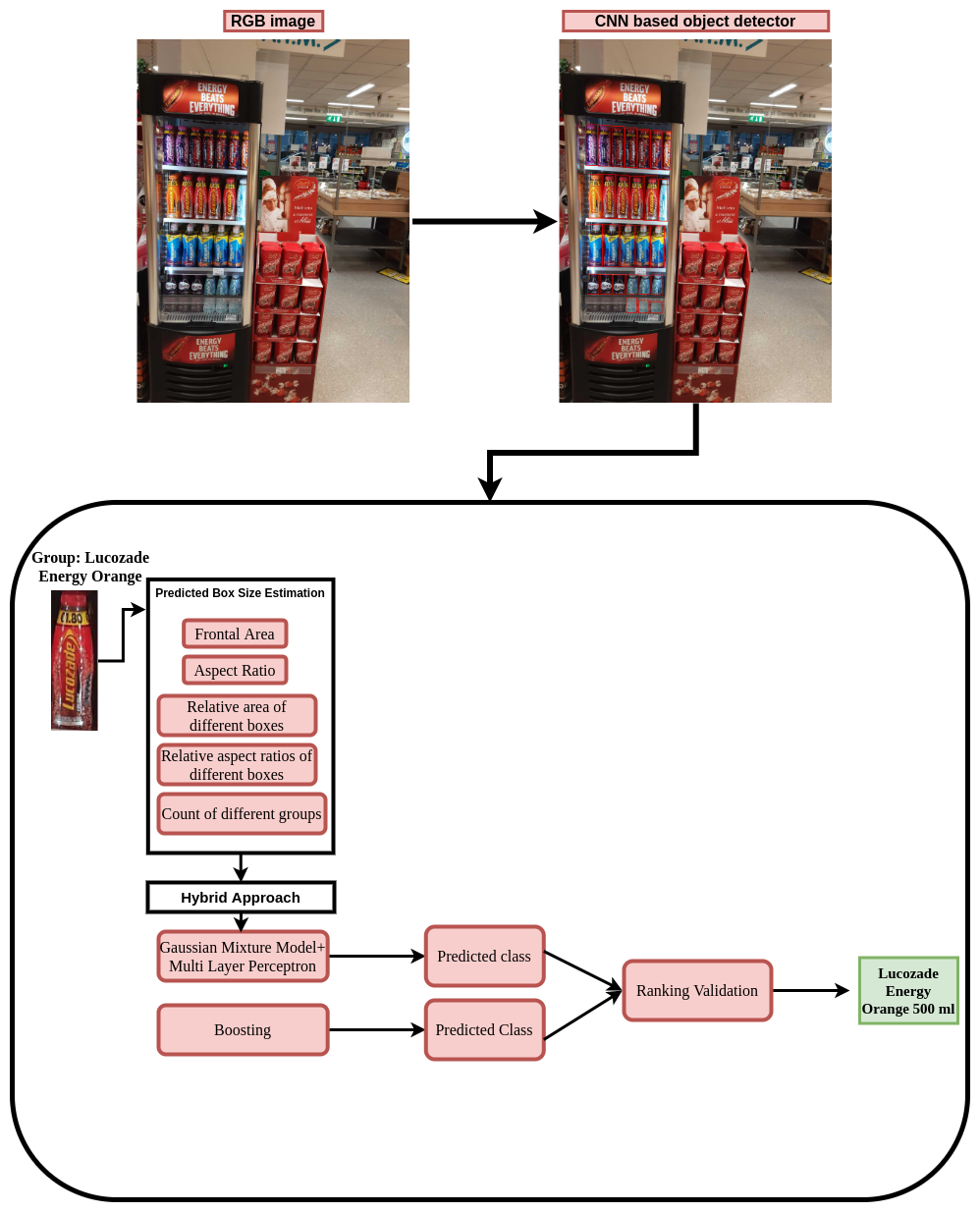}
  \caption{Our three steps modular approach for size reasoning from object detection to object classification and finally a hybrid approach to classify groups into their respective classes.}
  \label{fig:method}
\end{figure*}

\section{Methodology}

To determine different sized variants of products, we follow a modular approach which has three steps; 1) a CNN object detector to predict bounding boxes of products on a shelf image, 2) a CNN classifier to predict groups corresponding to those bounding boxes 3) size variants model to classify those groups into their respective classes. This work details the third step of the wrokflow. The knowledge of the brand we get from step 2 and relative sizes and aspect ratios of products we get from step 1 is used for size reasoning in the third step. Figure ~\ref{fig:method} shows the three modules of workflow.

\subsection{CNN based object detector and classification}
We already have an object detector to extract retail products trained on our in-house datasets based on the state of the art approaches \cite{varadarajan2020benchmark}. This is the first step to extract bounding boxes around products from shelf images

CNNs can be used to classify retail products with high accuracy. The products extracted by the object detector are passed through a CNN classifier to determine their group. \cite{MayankSrivastavaBagClassification}

\subsection{Modelling for Size based Reasoning}
The CNN object detection and classification models predict bounding boxes and groups respectively in a shelf image. We use these predictions to generate features for size variant classification model. Due to retail shelf photos being taken at different scales and distances from the shelf, no absolute measurement(length, width, area of products in pixels and square pixels respectively) would contain any useful information. We have to depend on relative measurements, like relative area of a box with respect to another box of the common or different group from the same image. The features we use are relative aspect ratio of boxes and relative frontal area [length X width of box being classified divided by length X width of other boxes in the image].

The size based reasoning problem can be understood as a classification problem where a product's bounding box [measurements in pixel] and its group [brand determined by CNN classifier] is available and we need to classify each product instance (henceforth called a size analysis candidate) into one of the possible size variants of the group/brand [classes in our work's terminology]. Thus we need to train multiple classifiers, one for each group to determine the box as one of the classes belonging to that group.

A size analysis candidate will have the following features in an image when being classified:
1. Aspect Ratio with respect to all other boxes in the image, each belonging to same group as the candidate or some other group.
2. Relative frontal area with respect to all other boxes in the image, belonging to same or other groups.
So the number of features while classifying a candidate is also not fixed and varies depends upon the number of products and number of groups in the image. Based on these features, we have to determine which class a candidate belongs to.

\begin{figure*}[tb]
  \centering
  \includegraphics[width=0.9\textwidth]{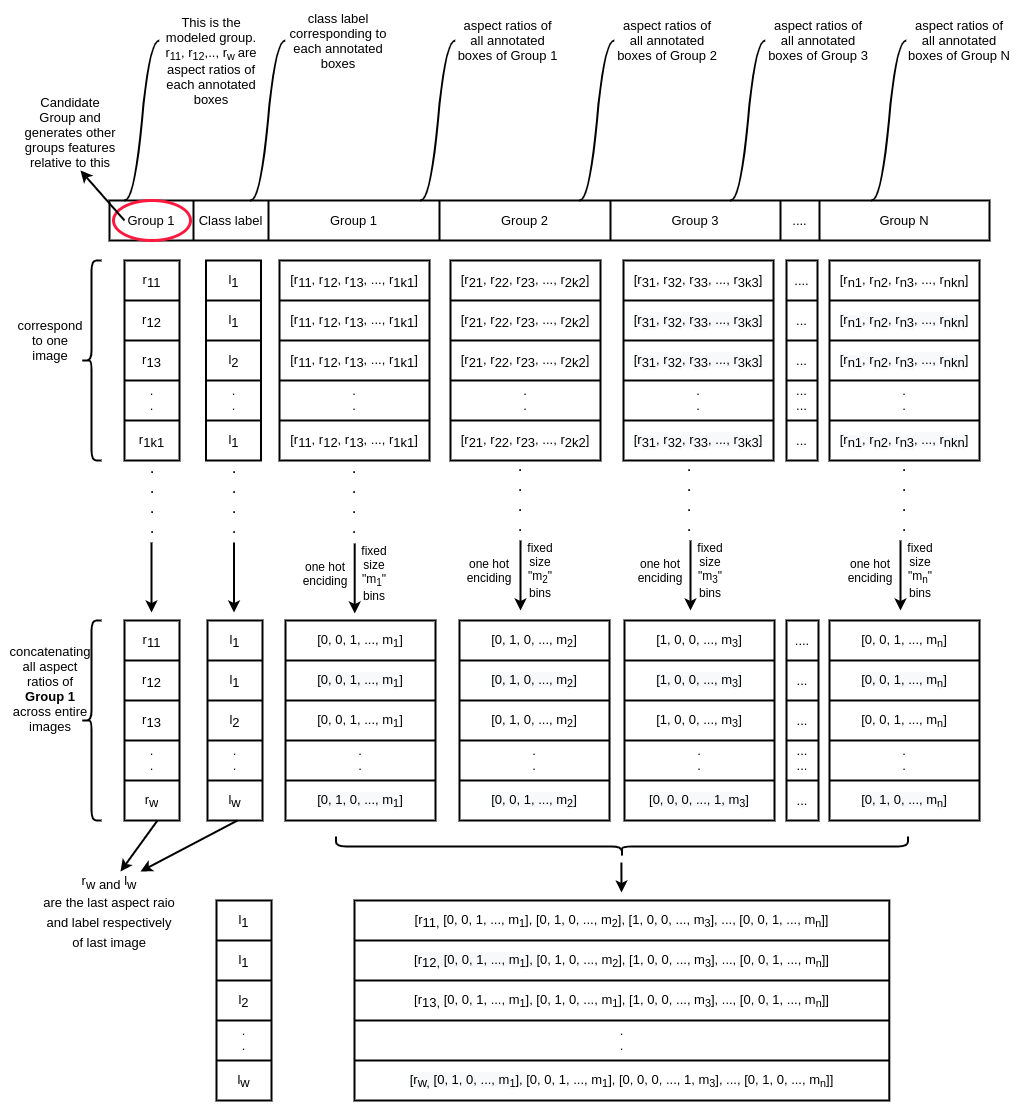}
  \caption{The features generation procedure for the boosting approach.}
  \label{fig:boosting}
\end{figure*}

\subsubsection{Common Features Pipeline} 
There can be multiple boxes the object detection algorithm determines in an image. The size dimension of the boxes are very much depend on how the image is captured. Like if an image is taken from far away, then size of each boxes are relatively smaller compared to when the same image is taken very closely. Due to this reason, we can only use relative area of boxes and aspect ratio of boxes as features in any downstream tasks. We also know the group of each of these boxes as determined by the CNN. So the final task is to classify a box [size analysis candidate] into one of the size variants of the group it belongs to, given its aspect ratio and relative area to boxes of various groups in an image the candidate comes from. Since each image has a different combination of products, we would have different number of determined area ratios for size analysis candidates. To explain using symbols a size analysis candidate $b_{i}$ is $[(area_{i},aspect-ratio_{i},group-label_{i}),class-label_{i}]$ pair, which cannot be used with absolute number for area. Thus, we take area feature of $b_{i}$ to be area ratio with other boxes $[ ( area_{k} / area_{i}, group-label_{k} , group-label_{i} ) , (area_{k+1} / area_{i}, group-label_{k+1}, group-label_{i}) ... (area_{k+m} / area_{i}, group-label_{k+m}, group-label_{i} ) ]$ where $m+1$ size analysis candidates [ $k..k+m$ ] / boxes are detected in the image from where $b_{i}$ was taken. Here m can vary, thus giving a vector of variable dimensions. Let's simplify notation for $b_{i}$ saying $r_{i}$ is aspect ratio of the size analysis candidate, ${a_{i}}$ is the its area, both determined from object detector output, $g_{i}$ is its group label determined from CNN classifier output, $c_{i}$ is the output class or the size variant when we are training. The final training vector for $b_{i}$ can thus denoted by pair $[ ( r_{i}, g_{i}, a_{k}/a_{i}, g_{k}, a_{k+1}/a_{i}, g_{k+1} .. a_{k+m}/a_{i}, g_{k+m} ) , c_{i} ]$ . Beyond this point, both the methods take different approaches to processing features.

The two methods we propose have different ways to process features. In the first method, varying dimensions of features are coalesced into a fixed sized feature vector. In the second method, we feed the varying length features we receive into a Neural Network that can train on variable number of features.

\subsubsection{XGBOOST specific features}
Discussing feature generation for XGBOOST classification method [referred to as method 1 henceforth] In the diagram shown in Fig. ~\ref{fig:boosting}, for each size analysis candidate box $b_{i}$, we make a vectors of all other boxes' [from the same image as $b_{i}$] aspect ratios collected according to group labels. Say $r_{11}, r_{12}, ... r_{1k_1}$ is vector of aspect ratios of Group Label 1 present in an image. Besides these, there can be other groups e.g., Group 2, Group 3, ..., Group N also present in that same image. We make vectors of aspect ratios of all boxes present in the image using their respective group labels. Like $r_{21}, r_{22}, ... r_{2k_2}$ for Group 2, $r_{31}, r_{32}, ... r_{3k_3}$ for Group 3 and so on. Since we want to train classifier for 1 Group Label as one model [classifying a brand into one of the size variants], we perform such vector creation for all boxes across all images. So for example, you have $r_{11}, r_{12}, ... r_{1k_1}$ from first image, $r_{11}, r_{12}, ... r_{1k_1}$ from second image and so forth when you want to train a classifier to recognize candidates from group label 1. Now aspect ratio part of a vector for a size analysis candidate box q, from image l with group label p starts looking like $(r_{pq},[r_{11}, r_{12} .. ]_{l},[r_{21}, r_{22} .. ]_{l}..[r_{p1}, r_{p2} .. ]_{l}..),c_{q}$. We also create similar vector of relative areas, that is, $a_{j}/a_{i}, g_{j}$ values and add them on to the vector of a candidate box $b_{i}$. \\*

In method 1, as shown in Fig  ~\ref{fig:boosting}, for each Group, we perform fixed size binning across the columns. As columns of other groups contains variable number boxes so to make them fixed size, we use Discretization (or binning) which is similar to constructing histograms for continuous data. It is a way to partition continuous features into discrete or fixed-size values which are one-hot encoded. One-hot encoded discretized features can make a model more expressive, while maintaining interpretability. The fixed size one-hot features and corresponding labels are used in the training of Boosting model. However, by converting the continuous data into a fixed size bins, there can be loss of essential information about the distribution of size of boxes and may lose some latent features too.

\subsubsection{Neural Network specific GMM features}
\begin{figure*}[tb]
  \centering
  \includegraphics[width=0.9\textwidth]{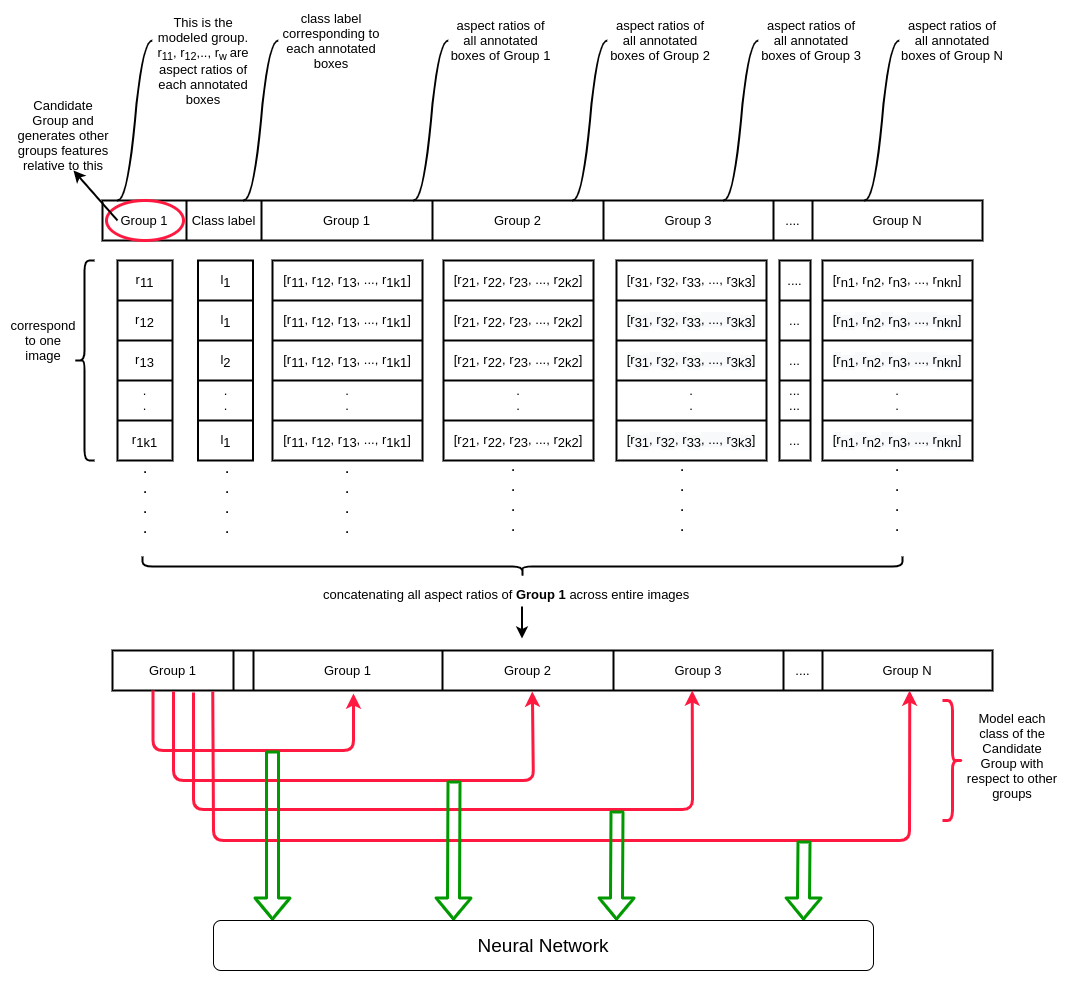}
  \caption{The features generation procedure for the bayesian approach.}
  \label{fig:bayesian}
\end{figure*}

In method 2 or Neural Network method, we use Gaussian Mixture Models to create features and then pass the features to a Neural Network that can take variable length input. On the train set, we model a set of GMM for each class label $c^j$. We create vectors from train data of form $[c^j{_{i}},g_{i},((r_{i}),(a_{k}/a_{i}),g_{k})]$ for all values $j^{th}$ class. This is vector of all candidate boxes belonging to a specific class, along with their aspect ratio, group label, area ratio with respect to candidates of other group labels. Given we know all possible sized variants of the available groups, we know that for each combination of a class label, its group label and any other group can have fixed number of maximum possible gaussians that is number of class labels for the other group in combination. To understand in symbols a tuple  $c^j,g_{c^j},gw$ can have at max $u$ combinations, where u is the number of different sized variants in the brand $gw$. Here $gw$ is a group whose boxes/candidates occur in same image as a candidate of class $c^j$. The maximum number of gaussians is independent of number of size variants of the brand $g_{c^j}$ corresponding to $c^j$. Effectively we try to model the area ratio of class $c^j$ with classes of $gw$ as gaussians when we model $c^j,gw$ as a gaussian mixture. So, data belonging to each combination of $c^j,gw$ is modelled as a 2D gaussian mixture of $(r,a_{gw}/a_{c^j})$ values. Now given we have a tuple of type $(r,a_{gw}/a_{.})$ like that at time of inference, we can calculate $p(c^j | (r,a_{gw}/a_{.}) , gmm(r,a_{gw}/a_{c^j}))$. The feature vector thus passed to the Neural Network for training is $[(r_{i},a_{g^w_{1}/a_{i}},p(c^v)_{1},p(c^{v+1})_{1},p(c^{v+2})_{1},..p(c^{v+n})_{1}..a_{g^w_{y}/a_{i}},p(c^v)_{y}..p(c^{v+n})_{y}),c_{i}]$. As the reader can make out, the feature length is variable depending upon number of boxes/candidates in the image $y$. Please note $c^v$ through $c^{v+n}$ are the classes [size level variants] of group [brand] $g_{i}$ of the candidate/box being classified. $g^w$ is the group label of any other candidate in the image. Figure ~\ref{fig:bayesian} shows the feature generation process in contrast with the fixed sized features created for XGBOOST. While calculating GMM for each $c^j$ by Bayesian method, the final values of $r$ and $a_{gw}/a_{c^j}$.


% ...

\subsection{Size variants classification}
We follow two different approaches for our size variants classification. First method uses the XGBoost algorithm directly on the features obtained, while second method uses GMM to further process features which are used as input to a neural network. 

\subsubsection{Gradient Boosting Model}
For each Group, we create the features of aspect ratio and area ratio as discussed above. However, each image need not contain all possible groups. So, we get blank value corresponding to group which is not available in the particular image. By merging all images which contain a Group we get lots of null values corresponding to the other groups which are not present in the images. For example, let say if there are n products in a store and a particular image contains n/3 products only then, the corresponding features of 2n/3 products which are not present in the image will be zero. Features created by groups co-occurring less than a  value are not used in processing. We evaluate multiple Machine Learning models on this fixed length feature vector and XGBOOST turns out to be giving best results. The final set of models have one model corresponding to each group, classifying a size analysis candidate as one possible class label for the group the classifier is trained on. The upstream CNN classifier makes the group label available for each box before the XGBOOST feature extraction and inference is trained or tested.

\subsubsection{Neural Network}
The GMM based feature generation gives a dense matrices of varying dimensions for each size analysis candidate. Like XGBOOST, each Group Label has a separate model to classify it into one of the classes. Each  $p(c^j | (r,a_{.}/a_{gw})$ set of features is passed through a dense layer specific to the combination to obtain an representation and all such representations for a size analysis candidate box are averaged to be passed through a dense layer which classifies the box into one of the size variants.

\section{Result}

\begin{figure*}[tb]
  \centering
  \includegraphics[width=\textwidth]{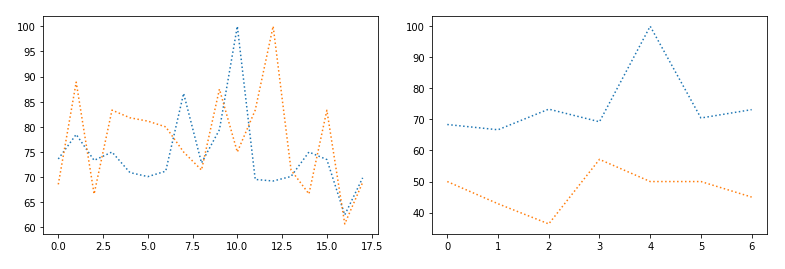}
  \caption{}
  \label{fig:d1}
\end{figure*}

The accuracy as classification accuracy for each group level model. Graphs shown in Figure \ref{fig:d1} show $class number$ on the X axis and the accuracy of both methods on each of the groups. Both the shown graphs show performance on a different set of group labels. Orange line represents accuracies of XGBOOST models while blue line represents Neural Network model. The models are competitive in the left graph, but Neural Network method works better in the right graph.

When analyzed for reasons of difference/similarity in performance, we found that if a group contains a high percentage of noisy and ill detected annotation boxes, the Neural Network based approach performed significantly well compared to the boosting based methods as shown in the Figure \ref{fig:d1}. However, if annotation boxes of a group follow good size differences among classes and fewer errors in box dimensions, both boosting and Neural Network methods are comparable (left figure).

\section{Conclusion}
We propose two Machine Learning methods to do size based inference for objects on retail shelf. Complex feature engineering is needed for models to perform such deductions. Neural Network method's feature engineering takes noise into account by modelling GMM features within a hierarchial model accounting for noise.

{\small
\bibliographystyle{ieee_fullname}
\bibliography{cvpr}

\begin{thebibliography}{1}\itemsep=-1pt

\bibitem{Chen2016XGBoost:System}
Tianqi Chen and Carlos Guestrin.
\newblock {XGBoost: A scalable tree boosting system}.
\newblock In {\em Proceedings of the ACM SIGKDD International Conference on
  Knowledge Discovery and Data Mining}, volume 13-17-August-2016, pages
  785--794. Association for Computing Machinery, 8 2016.

\bibitem{he2017mask}
Kaiming He, Georgia Gkioxari, Piotr Doll{\'a}r, and Ross Girshick.
\newblock Mask r-cnn.
\newblock In {\em Proceedings of the IEEE international conference on computer
  vision}, pages 2961--2969, 2017.

\bibitem{Hou2008RobustData}
Shaobo Hou and Aphrodite Galata.
\newblock {Robust estimation of gaussian mixtures from noisy input data}.
\newblock In {\em 26th IEEE Conference on Computer Vision and Pattern
  Recognition, CVPR}, 2008.

\bibitem{KrizhevskyImageNetNetworks}
Alex Krizhevsky, Ilya Sutskever, and Geoffrey~E Hinton.
\newblock {ImageNet Classification with Deep Convolutional Neural Networks}.
\newblock Technical report.

\bibitem{MayankSrivastavaBagClassification}
Muktabh Mayank~Srivastava.
\newblock {Bag of Tricks for Retail Product Image Classification}.
\newblock Technical report.

\bibitem{Simonyan2014VeryRecognition}
Karen Simonyan and Andrew Zisserman.
\newblock {Very Deep Convolutional Networks for Large-Scale Image Recognition}.
\newblock 9 2014.

\bibitem{varadarajan2020benchmark}
Srikrishna Varadarajan, Sonaal Kant, and Muktabh~Mayank Srivastava.
\newblock Benchmark for generic product detection: a low data baseline for
  dense object detection.
\newblock In {\em International Conference on Image Analysis and Recognition},
  pages 30--41. Springer, 2020.

\bibitem{Wan2019AClassification}
Huan Wan, Hui Wang, Bryan Scotney, and Jun Liu.
\newblock {A novel gaussian mixture model for classification}.
\newblock In {\em Conference Proceedings - IEEE International Conference on
  Systems, Man and Cybernetics}, volume 2019-October, pages 3298--3303.
  Institute of Electrical and Electronics Engineers Inc., 10 2019.

\bibitem{ZongDEEPDETECTION}
Bo Zong, Qi Song, Martin Renqiang~Min, Wei Cheng, Cristian Lumezanu, Daeki Cho,
  and Haifeng Chen.
\newblock {DEEP AUTOENCODING GAUSSIAN MIXTURE MODEL FOR UNSUPERVISED ANOMALY
  DETECTION}.
\newblock Technical report.

\end{thebibliography}
}

\end{document}